# Copula-based synthetic data augmentation for machine-learning emulators

David Meyer[1,2] (ORCID: 0000-0002-7071-7547)

Thomas Nagler[3] (ORCID: 0000-0003-1855-0046)

Robin J. Hogan[4,1] (ORCID: 0000-0002-3180-5157)

[1]Department of Meteorology, University of Reading, Reading, UK

[2]Department of Civil and Environmental Engineering, Imperial College London, London, UK

[3]Mathematical Institute, Leiden University, Leiden, The Netherlands

[4]European Centre for Medium-Range Weather Forecasts, Reading, UK

Correspondence to David Meyer (d.meyer@pgr.reading.ac.uk)

**Abstract**

Can we improve machine-learning (ML) emulators with synthetic data? If data are scarce or expensive to source and a physical model is available, statistically generated data may be useful for augmenting training sets cheaply. Here we explore the use of copula-based models for generating synthetically augmented datasets in weather and climate by testing the method on a toy physical model of downwelling longwave radiation and corresponding neural network emulator. Results show that for copula-augmented datasets, predictions are improved by up to 62 % for the mean absolute error (from 1.17 to 0.44 W m$^{-2}$).

1. Introduction

The use of machine learning (ML) in weather and climate is becoming increasingly popular (Huntingford et al., 2019; Reichstein et al., 2019). ML approaches are being applied to an increasingly diverse range of problems for improving the modelling of radiation (e.g. Cheruy et al., 1996; Chevallier et al., 1998, 2000; Krasnopolsky et al., 2005; Meyer et al., 2021; Ukkonen et al., 2020; Veerman et al., 2021), ocean (e.g. Bolton and Zanna, 2019; Krasnopolsky et al., 2005), chemistry (e.g. Nowack et al., 2018), and convection (e.g. Krasnopolsky et al., 2013), as well as the representation of sub-grid processes (e.g. Brenowitz and Bretherton, 2018; Gentine et al., 2018; O'Gorman and Dwyer, 2018; Rasp et al., 2018), and the post-processing of model outputs (e.g. Krasnopolsky and Lin, 2012; Rasp and Lerch, 2018).

When it comes to training ML models for weather and climate applications two main strategies may be identified: one in which input and output pairs are directly provided (e.g. both come from observations) and a second in which inputs are provided but corresponding outputs are generated through a *physical model* (e.g. parameterization schemes or even a whole weather and climate model). Although the former may be considered the most common training strategy in use today, when the underlying physical processes are well understood (e.g. radiative transfer) and numerical codes are available, the latter may be of particular interest for developing one-to-one *emulators* (i.e. statistical surrogates of their physical counterparts), which can be used to improve computational performance for a trade-off in accuracy (e.g. Chevallier et al., 1998; Meyer et al., 2021; Ukkonen et al., 2020; Veerman et al., 2021). Here, for clarity, we will only be focusing on the latter case and refer to them as emulators.

In ML, the best way to make a model more generalizable is to train it on more data (Goodfellow et al., 2016). However, depending on the specific field and application, input data may be scarce, representative of only a subset of situations and domains, or, in the case of synthetically generated data, require large computational resources, bespoke infrastructures, and specific domain knowledge. For example, generating atmospheric profiles using a general circulation model (GCM) may require in-depth knowledge of the GCM and large computational resources (e.g. data used in Meyer et al., 2021).

A possible solution to these issues may be found by augmenting the available input dataset with more samples. Although this may be a straightforward task for classification problems (e.g. by translating or adding noise to an image), this may not be the case for parameterizations of physical processes used in weather and climate models. In this context, it is common to work with high-dimensional and strongly dependent data (e.g. between physical quantities such as air temperature, humidity, and pressure across grid points). Although this dependence may be well approximated by simple physical laws (e.g. the ideal gas law for conditions found in the Earth's atmosphere), the generation of representative data across multiple dimensions for most weather and climate applications is challenging (e.g. the nonlinear relationship between cloud properties, humidity, and temperature).

To serve a similar purpose as real data, synthetically generated data thus need to preserve the statistical properties of real data in terms of individual behaviour and (inter-)dependences. Several methods may be suitable for generating synthetic data such as copulas (e.g. Patki et al., 2016), variational autoencoders (e.g. Wan et al., 2017), and, more recently, generative adversarial networks (GANs; e.g. Xu and Veeramachaneni, 2018). Although the use of GANs for data generation is becoming increasingly popular among the core ML community, these require multiple models to be trained, leading to difficulties and computational burden (Tagasovska et al., 2019). Variational approaches, on the other hand, make strong distributional assumptions that are potentially detrimental to generative models (Tagasovska et al., 2019). Compared to black-box deep-learning models, the training of vine copulas is relatively easy and robust, while taking away a lot of guesswork in specifying hyperparameters and network architecture. Furthermore, copula models give a direct representation of statistical distributions, making them easier to interpret and tweak after training. As such, copula-based models have been shown to be effective for generating synthetic data comparable to real data in the context of privacy protection (Patki et al., 2016).

The goal of this paper is to improve ML emulators by augmenting the physical model's inputs using copulas. We give a brief overview of methods in Sect. 2.1 with specific implementation details in Sect. 2.2–2.5. Results are shown in Sect. 3, with a focus on evaluating synthetically generated data in Sect. 3.1 and ML predictions in Sect. 3.2. We conclude with a discussion and prospects for future research in Sect. 4.

## 2. Material and methods

### 2.1 Overview

The general method for *training* an ML emulator for a set of $N$ samples involves the use of paired *inputs* $\boldsymbol{x} = \{x_1, \dots, x_N\}$ and *outputs* $\boldsymbol{y} = \{y_1, \dots, y_N\}$ to find the best function approximation for a specific architecture and configuration. For *inference*, the trained ML emulator is then used to predict new outputs $\boldsymbol{y}^*$ from inputs $\boldsymbol{x}^*$. Outputs $\boldsymbol{y}$ are generated through a physical model from $\boldsymbol{x}$ and fed to the ML emulator for training (Fig. 1a). In this paper we introduce an additional step: augmentation through copula-based synthetic data generation (Fig. 1b). The method is demonstrated with a toy model of downwelling radiation as the physical model (Sect. 2.4) and a simple feed-forward neural network (FNN) as the ML emulator (Sect. 2.5). To evaluate the impact of copula-generated synthetic data on predictions we focus on predicting vertical profiles of longwave radiation from those of dry-bulb air temperature, atmospheric pressure, and cloud optical depth (other parameters affecting longwave radiative transfer, such as gas optical depth, are treated as constant in the simple model described in Sect. 2.4). This task is chosen at it allows us to (i) evaluate copula-based models for generating correlated multidimensional data (e.g. with dependence across several quantities and grid points), some of which (e.g. cloud optical depth) are highly non-Gaussian; (ii) develop a simple and fast toy physical model that may be representative of other physical parameterizations such as radiation, (urban) land surface, cloud, or convection schemes; and (iii) develop a fast and simple ML emulator used to compute representative statistics. Here we define case (a) as the *baseline* and generate six different subcases for case (b) using (i) three levels of data *augmentation factors* (i.e. either 1x, 5x, or 10x the number of profiles in the real dataset) (ii) generated from three different copula types. In the following sections we give background information and specific implementation details about the general method used for setting up the source data (Sect. 2.2), data generation (Sect. 2.3), target generation (Sect. 2.4), and estimator training (Sect. 2.5) as shown in Fig. 1b.

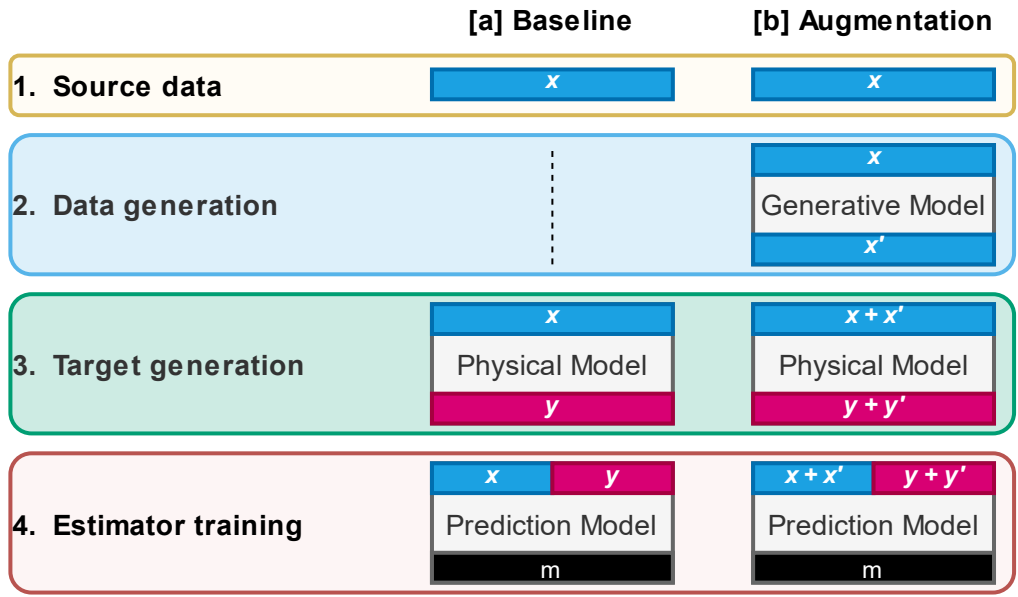

**Figure 1.** General strategies identified for training ML emulators. (**a**) Inputs $x$ are fed to the physical model to generate corresponding outputs $y$; $x$ and $y$ are used to train the ML emulator. (**b**) A data generation model (here copula) is fitted to inputs $x$ to generate synthetic inputs $x'$; inputs $x$ and $x'$ are fed to the physical model to generate corresponding outputs $y$ and $y'$; both $x$, $x'$ and $y$, $y'$ are used to train the ML emulator. After training, the model ($m$; e.g. architecture and weights) is saved and used for inference on new data.

**2.2 Source Data**

Inputs are derived from the EUMETSAT Numerical Weather Prediction Satellite Application Facility (NWP-SAF; Eresmaa and McNally, 2014) dataset. This contains a representative collection of 25 000 atmospheric profiles previously used to evaluate the performance of radiation models (e.g. Hocking et al., 2021; Hogan and Matricardi, 2020). Profiles were derived from 137-vertical-level global operational short-range ECMWF forecasts correlated in more than one dimension (between quantities and spatially across levels) and extending from the top of the atmosphere (TOA; 0.01 hPa; level 1) to the surface (bottom of the atmosphere; BOA; level 137). Inputs consist of profiles of dry-bulb air temperature ($T$ in K; Figure 2a), atmospheric pressure ($p$ in hPa; Figure 2b), and cloud layer optical depth ($\tau_c$; Figure 2c). $\tau_c$ is derived from other quantities to simplify the development of models as described in Sect. 2.4. Dry-bulb air temperature, atmospheric pressure, and cloud layer optical depth are then used as inputs to the physical model (Sect. 2.4) to compute outputs containing profiles of downwelling longwave radiation ($L^{\downarrow}$ in W m$^{-2}$; Figure 2d). As both copula models and ML emulator work on two-dimensional data, data are reshaped to input **X** and output **Y** matrices with each profile as row (sample) and flattened level and quantity as column (feature) and reconstructed to their original shape where required. Prior to being used, source data are shuffled at random and split into three batches of 10 000 profiles (40 %) for training (**X**$_{\text{train}}$, **Y**$_{\text{train}}$), 5000 (20 %) for validation (**X**$_{\text{val}}$, **Y**$_{\text{val}}$), and 10 000 (40 %) for testing (**X**$_{\text{test}}$, **Y**$_{\text{test}}$).

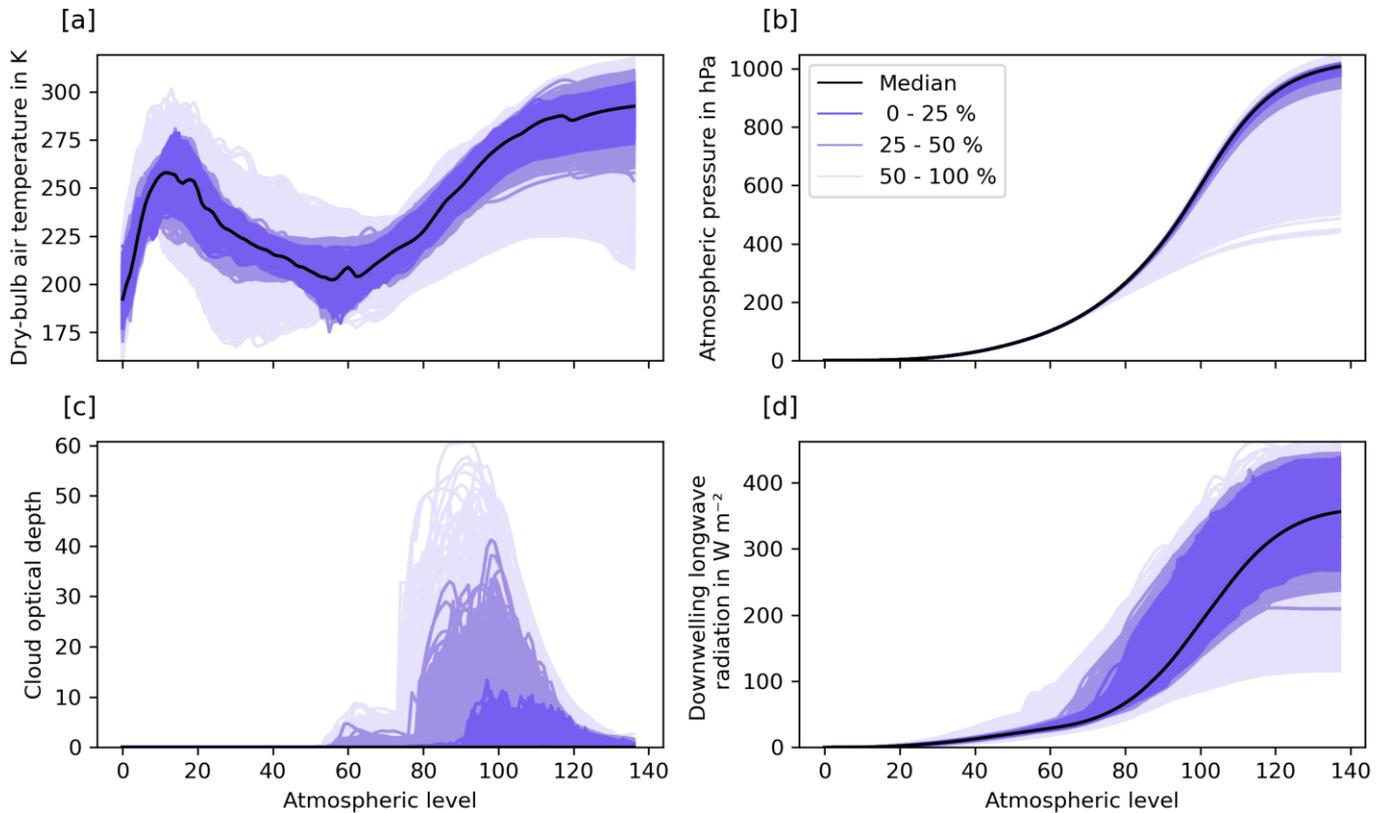

**Figure 2.** Profiles of (**a**) dry-bulb air temperature, (**b**) atmospheric pressure, and (**c**) cloud layer optical depth from the NWP-SAF dataset (25 000 profiles; Eresmaa and McNally, 2014) as well as (**d**) corresponding profiles of longwave radiation computed using the toy physical model described in Sect. 2.4. Profiles are ordered using band depth statistics (López-Pintado and Romo, 2009), shown for their most central (median) profile, and grouped for the central 0 %–25 %, 25 %–50 %, and 50 %–100 %.

**Table 1.** Profiles of input and output quantities used in this study. Input quantities are dry-bulb air temperature $T$, atmospheric temperature $p$, and cloud layer optical depth $\tau_c$. $T$ and $p$ are taken directly from the NWP-SAF dataset (Eresmaa and McNally, 2014), and $\tau_c$ is derived from other quantities as described in Sect. 2.4. The output quantity downwelling longwave radiation $L^\downarrow$ is computed using the physical model described in Sect. 2.4. Atmospheric model levels are 137 for full levels (FLs) and 138 for half-levels (HLs).

| Symbol | Name | Unit | Dimension |
|---|---|---|---|
| Inputs | | | |
| $T$ | Dry-bulb air temperature | K | FL |
| $p$ | Atmospheric pressure | Pa | FL |
| $\tau_c$ | Cloud optical depth | 1 | FL |
| Output | | | |
| $L^\downarrow$ | Downwelling longwave radiation | W m$^{-2}$ | HL |

## 2.3 Data generation

Data generation is used to generate additional input samples (here atmospheric profiles) to be fed to the physical model (Sect. 2.4) and ML (Sect. 2.5) emulator. Optimally, synthetically generated data should resemble the observed data as closely as possible with respect to (i) the individual behaviour of variables (e.g. the dry-bulb air temperature at a specific level) and (ii) the dependence across variables and dimensions (e.g. the dry-bulb air temperature across two levels). Copulas are statistical models that allow these two aims to be disentangled (Trivedi & Zimmer, 2006; Joe, 2014) and to generate new samples that are statistically similar to the original data in terms of their individual behaviour and dependence.

### 2.3.1 Background on copula models

Suppose we want to generate synthetic data from a probabilistic model for $n$ variables $Z_1, \ldots, Z_n$. To achieve the first aim, we need to find appropriate *marginal cumulative distributions* $F, \ldots, F_n$. A simple approach is to approximate them by the corresponding empirical distribution functions. To achieve the second aim, however, we need to build a model for the *joint distribution function* $F(z_1, ..., z_n)$. The key result, Sklar's theorem (Sklar, 1959), states that any joint distribution function can be written as

$$F(z_1, ..., z_n) = C(F_1(z_1), ..., F_n(z_n)). \quad (1)$$

The function $C$ is called the copula and encodes the dependence between variables.

Copulas are distribution functions themselves. More precisely, if all variables are continuous, $C$ is the joint distribution of the variables $U_1 = F_1(Z_1), ..., U_n = F_n(Z_n)$. This fact facilitates estimation and simulation from the model. To estimate the copula function $C$, we (i) estimate marginal distributions $\widehat{F}_1, ..., \widehat{F}_n$, (ii) construct *pseudo-observations* $\widehat{U}_1 = \widehat{F}_1(Z_1), ..., \widehat{U}_n = \widehat{F}_n(Z_n)$, and (iii) estimate $C$ from the pseudo-observations. Then, given estimated models $\widehat{C}$ and $\widehat{F}_1, ..., \widehat{F}_n$ for the copula and marginal distributions, we can generate synthetic data as follows.

1. Simulate random variables $U_1, ..., U_n$ from the estimated copula $\widehat{C}$.
2. Define $Z_1 = \widehat{F}_1^{-1}(X_1), ..., Z_n = \widehat{F}_n^{-1}(X_n)$.

**2.3.2 Parametric copula families**

In practice, it is common to only consider sub-families of copulas that are conveniently parameterized. There is a variety of such parametric copula families. Such families can be derived from existing models for multivariate distributions by inverting the equation of Sklar's theorem:

$$C(u_1, ..., u_n) = F(F_1^{-1}(u_1), ..., F_n^{-1}(u_n)). \quad (2)$$

For example, we can take $F$ as the joint distribution function of a multivariate Gaussian and $F_1, ..., F_n$ as the corresponding marginal distributions. Then Eq. (2) yields a model for the copula called the Gaussian copula, which is parameterized by a correlation matrix. The Gaussian copula model includes all possible dependence structure in a multivariate Gaussian distribution. The benefit comes from the fact that we can combine a given copula with any type of marginal distribution, not just the ones the copula was derived from. That way, we can build flexible models with arbitrary marginal distributions and Gaussian-like dependence. The same principle applies to other multivariate distributions and many copula models have been derived, most prominently the Student's *t* copula and Archimedean families. A comprehensive list can be found in Joe (2014).

**2.3.3 Vine copula models**

When there are more than two variables ($n > 2$) the type of dependence structure these models can generate is rather limited. Gaussian and Student copulas only allow for symmetric dependencies between variables. Quite often, dependence is asymmetric, however. For example, dependence between $Z_1$ and $Z_2$ may be stronger when both variables take large values. Many Archimedean families allow for such asymmetries but require all pairs of variables to have the same type and strength of dependence.

Vine copula models (Aas et al., 2009; Czado, 2019) are a popular solution to this issue. The idea is to build a large dependence model from only two-dimensional building blocks. We can explain this with a simple example with just three variables: $Z_1, Z_2$, and $Z_3$. We can model the dependence between $Z_1$ and $Z_2$ by a two-dimensional copula $C_{1,2}$ and the dependence between $Z_2$ and $Z_3$ by another, possibly different, copula $C_{2,3}$. These two copulas already contain some information about the dependence between $Z_1$ and $Z_3$, the part of the dependence that is induced by $Z_2$. The missing piece is the dependence between $Z_1$ and $Z_3$ after the effect of $Z_2$ has been removed. Mathematically, this is the conditional dependence between $Z_1$ and $Z_3$ given $Z_2$ and can be modelled by yet another two-dimensional copula $C_{1,3|2}$. The principle is easily extended to an arbitrary number of variables $Z_1, \ldots, Z_n$. Algorithms for simulation and selection of the right conditioning order and parametric families for each (conditional) pair are given in Dißman et al. (2013).

Because all two-dimensional copulas can be specified independently, such models are extremely flexible and allow for highly heterogenous dependence structures. Using parametric models for pairwise dependencies remains a limiting factor, however. If necessary, it is also possible to use nonparametric models for the two-dimensional building blocks. Here, the joint distribution of pseudo-observations $(\widehat{U}_1, \widehat{U}_2)$ is estimated by a suitable kernel density estimator (see Nagler et al., 2017).

**2.3.4 Implementation**

Here we use Synthia (Meyer & Nagler, 2021) version 0.3.0 (Meyer & Nagler, 2020) with pyvinecopulib (Nagler & Vatter, 2020) to fit three different copula types: Gaussian, vine-parametric, and vine-nonparametric. Vine-parametric fits a parametric model for each pair in the model from the catalogue of Gaussian, Student, Clayton, Gumbel, Frank, Joe, BB1, BB6, BB7, and BB8 copula families and their rotations (see Joe, 2014, for details on these families) using the Akaike information criterion (AIC). Vine-nonparametric uses transformation local quadratic likelihood fitting as explained in Nagler et al. (2017). Copulas are fitted to $\mathbf{X}_{\text{train}}$ to generate synthetic training sets $\mathbf{X}'_{\text{train}}$ using three augmentation factors (i.e. each containing either 1x, 5x, or 10x the number of profiles in $\mathbf{X}_{\text{train}}$). $\mathbf{X}_{\text{train}}$ plus $\mathbf{X}'_{\text{train}}$ form augmented training sets containing 20 000 profiles (or double the amount of training data) for 1x augmentation factor and 60 000 and 110 000 profiles for 5x and 10x augmentation factors, respectively. As the generation of new profiles with copula models is random, the generation is also repeated 10 times for each case to allow meaningful statistics to be computed.

**2.4 Target generation**

Target generation is used to generate outputs from corresponding inputs using a physical model. Here, outputs are computed using a simple toy model based on Schwarzschild's equation (e.g. Petty, 2006) to estimate the downwelling longwave radiation under the assumption that atmospheric absorption does not vary with wavelength as

$$\frac{dL^\downarrow}{dz} = a(z)[B(z) - L^\downarrow], \tag{3}$$

where $z$ is the geometric height, $B$ is the Planck function at level $z$ (i.e. $B = \sigma_{\text{SB}} T^4$, where $\sigma_{\text{SB}}$ is the Stefan-Boltzmann constant; giving the flux in W m$^{-2}$ emitted from a horizontal black body surface), and $a$ is the rate at which radiation is intercepted and/or emitted. A common approximation is to treat longwave radiation travelling at all angles as if it were all travelling with a zenith angle of 53° (Elsasser, 1942): in this case $a = D\beta_e$, where $\beta_e$ is the extinction coefficient of the medium, and $D = 1/\cos(53) = 1.66$ is the diffusivity factor, which accounts for the fact that the effective path length of radiation passing through a layer of thickness $\Delta z$ is on average $1.66\Delta z$ due to the multiple different angles of propagation. In the context of ML, $a(z)$ and $B(z)$ are known and $F(z)$ is to be predicted. Here we use the difference in two atmospheric pressures expressed in sigma coordinates ($\Delta\sigma$, where $\sigma$ is the pressure $p$ at a particular height divided by the surface pressure $p_0$) instead of $z$. The layer optical depth $\tau = \beta_e \Delta z$ is calculated from the total-column gas optical depth $\tau_g$ and cloud layer optical depth $\tau_c$ as $\tau = \tau_c + \tau_g \Delta\sigma_i$, since $\Delta\sigma$ is the fraction of mass of the full atmospheric column in layer $i$. Then, as the downwelling flux at the top of the atmosphere is 0, the equation is discretized as follows assuming $B$ and $a$ are constant within a layer:

$$L^\downarrow_{i-1/2} = L^\downarrow_{i+1/2}(1 - \epsilon_i) + B_i \epsilon_i, \tag{4}$$

where $B_i$ is the Planck function of layer $i$, $\epsilon_i = 1 - e^{-a_i \Delta z} = 1 - e^{D\tau}$ is the emissivity of layer $i$, $L^\downarrow_{i+1/2}$ is the downwelling flux at the top of layer $i$, and $L^\downarrow_{i-1/2}$ is the downwelling flux at the bottom of layer $i$. We compute $L^\downarrow$ from $T$, $p$, and $\tau_c$ using the real

Published in Geoscientific Model Development. Please cite as: Meyer, D., Nagler, T., & Hogan, R. J. (2021). Copula-based synthetic data augmentation for machine-learning emulators. *Geoscientific Model Development*, *14*(8), 5205–5215. https://doi.org/10.5194/gmd-14-5205-2021NWP-SAF ($\mathbf{X}_\text{train}$) or augmented ($\mathbf{X}_\text{train}$ plus $\mathbf{X}'_\text{train}$) data. To reduce, and thus simplify, the number of quantities used in the physical model and ML emulator (Sect. 2.5), $\tau_c$ is pre-computed and used instead of vertical profiles of liquid and ice mixing ratios ($q_l$ and $q_l$) and effective radius ($r_l$ and $r_l$ in m) as $\frac{3}{2}\frac{\Delta p}{g}\left(\frac{q_l}{\rho_l r_l}+\frac{q_i}{\rho_i r_i}\right)$, where $\rho_l$ is the density of liquid water (1000 kg m$^{-3}$), $\rho_i$ is the density of ice (917 kg m$^{-3}$), $g$ is the standard gravitational acceleration (9.81 m s$^{-2}$). For $\tau_g$ we use a constant value of 1.7 determined by minimizing the absolute error between profiles computed with this simple model and the comprehensive atmospheric radiation scheme ecRad (Hogan & Bozzo, 2018).

**2.5 Estimator training**

As the goal of this paper is to determine whether the use of synthetic data improves the prediction of ML emulators, here we implement a simple feed-forward neural network (FNN). FNNs are one of the simplest and most common neural networks used in ML (Goodfellow et al., 2016) and have been previously used in similar weather and climate applications (e.g. Chevallier et al., 1998; Krasnopolsky et al., 2002). FNNs are composed of artificial neurons (conceptually derived from biological neurons) connected with each other; information moves forward from the input nodes through hidden nodes. The multilayer perceptron (MLP) is a type of FNN composed of at least three layers of nodes: an input layer, a hidden layer, and an output layer, with all but the input nodes using a nonlinear activation function.

Here we implement a simple MLP consisting of three hidden layers with 512 neurons each. This is implemented in TensorFlow (Abadi et al., 2016), and configured with the Exponential Linear Unit activation function, Adam optimizer, Huber loss, 1000-epoch limit, and early stopping with patience of 25 epochs. The MLP is trained with profiles of dry-bulb air temperature (Fig. 2a), atmospheric pressure (Fig. 2b), and layer cloud optical depth (Fig. 2c) as inputs and profiles of downwelling longwave radiation (Fig. 2d) as outputs. Inputs are normalized and both inputs and outputs are flattened into two-dimensional matrices as described in Sect. 2.2. The baseline case (Fig. 1a) uses 10 000 input profiles without data augmentation for training, and copula-based cases (Fig. 1b) use either 20 000, 60 000, or 110 000 profiles. The validation dataset $\mathbf{Y}_\text{val}$ of 5000 profiles is used as input for the early stopping mechanism, while the test dataset $\mathbf{Y}_\text{test}$ of 10 000 profiles is used to compute statistics (Sect. 3.2). Because of the stochastic nature of MLPs, training (and inference) is repeated 10 times for each case to allow meaningful statistics to be computed. Given that the generation of random profiles in the case of augmented datasets is also repeated 10 times (see Sect. 2.3.4), any case using data generation includes 100 iterations in total (i.e. for each data generation run, the estimator is trained 10 times).

**3 Results**

**3.1 Copula**

The quality of synthetic data is assessed in terms of summary statistics (e.g. Seitola et al., 2014) between the training $\mathbf{X}_\text{train}$ and copula-simulated $\mathbf{X}'_\text{train}$ datasets. For each copula type we compute a vector of summary statistics $\boldsymbol{s}_i = f(\boldsymbol{p}_i)$ where $f$ is the statistic function and $\boldsymbol{p}_i = \mathbf{D}\boldsymbol{w}$, with $\mathbf{D}$ a matrix of flattened source or simulated data and $\boldsymbol{w}$ a vector of random numbers for the $i$th iteration. Summary statistics are computed for mean, variance, and quantiles, iterating 100 times to allow meaningful statistics to be computed. As we consider random linear combinations of variables in source and copula-generated data, we expect these summaries to coincide only if both marginal distributions and dependence between variables are captured. Figure 3 shows scatterplots of summary statistics $\boldsymbol{s}_i$ for (a) mean, (b) variance, (c) standard deviation, and (d) 10 %, (e) 50 %, and (f) 90% quantiles. Real NWP-SAF data are shown on the $x$ axis and copula generated data on the $y$ axis, with each point corresponding to a random projection as described earlier (100 points in total total). For a perfect copula model, we expect all points to fall on the main diagonal, where $x = y$. Figure 3 shows that for all copula models, synthetically generated data are close to the real data, with larger errors in variance and standard deviation. Qualitatively, we can evaluate copula-generated profiles in terms of their overall shape and smoothness across multiple levels, as well as range and density at each level. To this end we plot a side-by-side comparison of source (Fig. 4, left panel) and Gaussian-copula-generated (Fig. 4, right panel) profiles showing the median profile and random selection of 90 profiles grouped in batches of 3 (i.e. each having 30 profiles) for the central 0 %–25 %, outer 25 %–50 %, and 50 %–100% quantiles calculated with band depth statistics (López-Pintado & Romo, 2009). Simulated profiles of dry-bulb air temperature (Fig. 4b) appear less smooth than the real ones across levels (Fig. 4a); however, both density and range are simulated well at each level. Simulated profiles of atmospheric pressure (Fig. 4d) are simulated well: they are smooth across all levels with similar range and density (Fig. 4c). The highly non-Gaussian and spiky profiles of cloud optical depth (Fig. 4e) make qualitative comparisons difficult; however, simulated profiles (Fig. 4f) have a similar range and density, with high density for low values, and most range between levels 80 and 120.

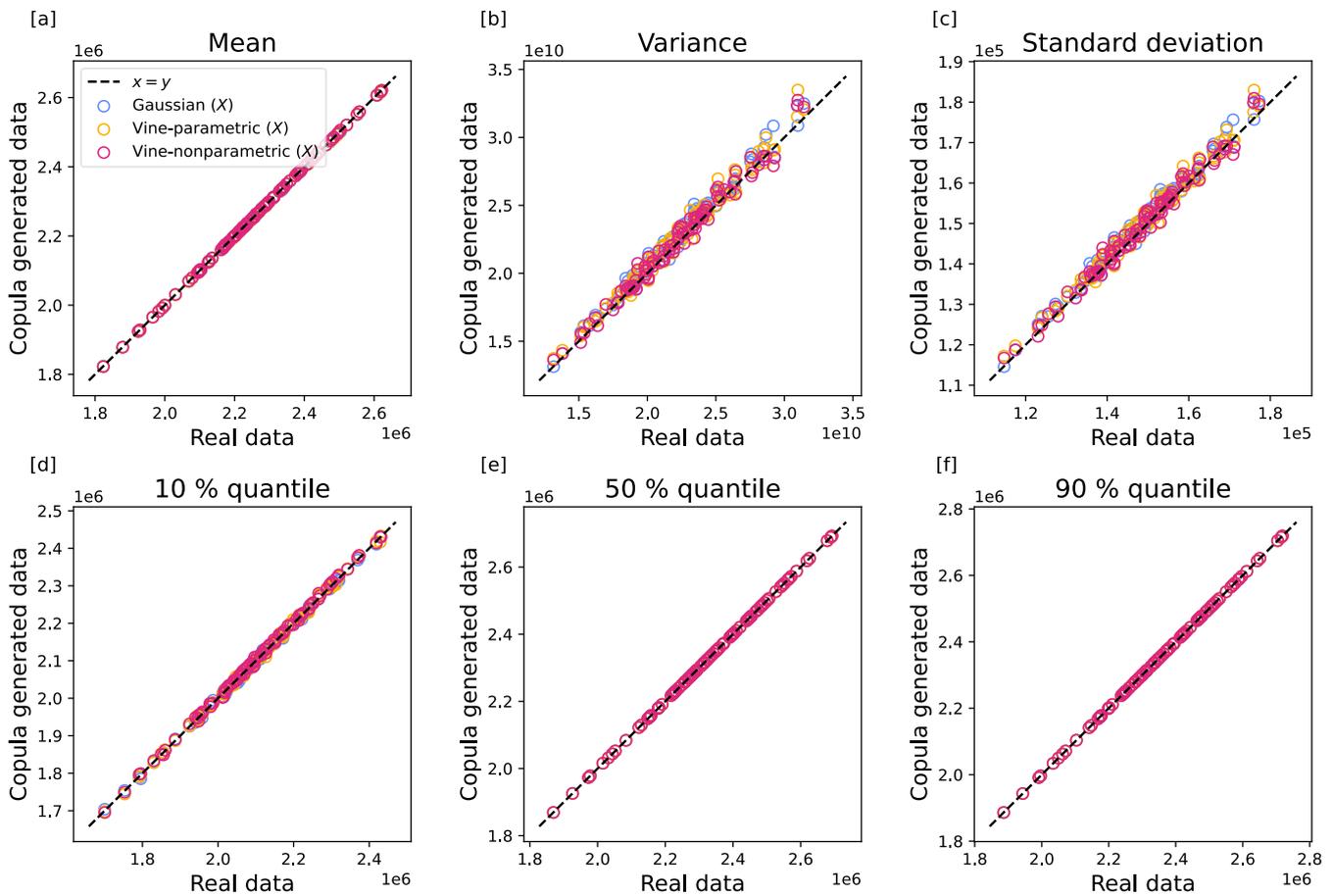

**Figure 3.** Summary statistics $s_i$ from 100 iterations for (**a**) mean, (**b**) variance, (**c**) standard deviation, and (**d**) 10 %, (**e**) 50 %, and (**f**) 90% quantiles. Each point corresponds to a statistic for a single iteration in arbitrary units. The $x$ axis represents the projection of real NWP-SAF $\mathbf{X_{train}}$, while the $y$ axis represents that of the copula-generated data $\mathbf{X'_{train}}$. Results are reported for Gaussian, vine-parametric, and vine-nonparametric copulas (see legend for keys).

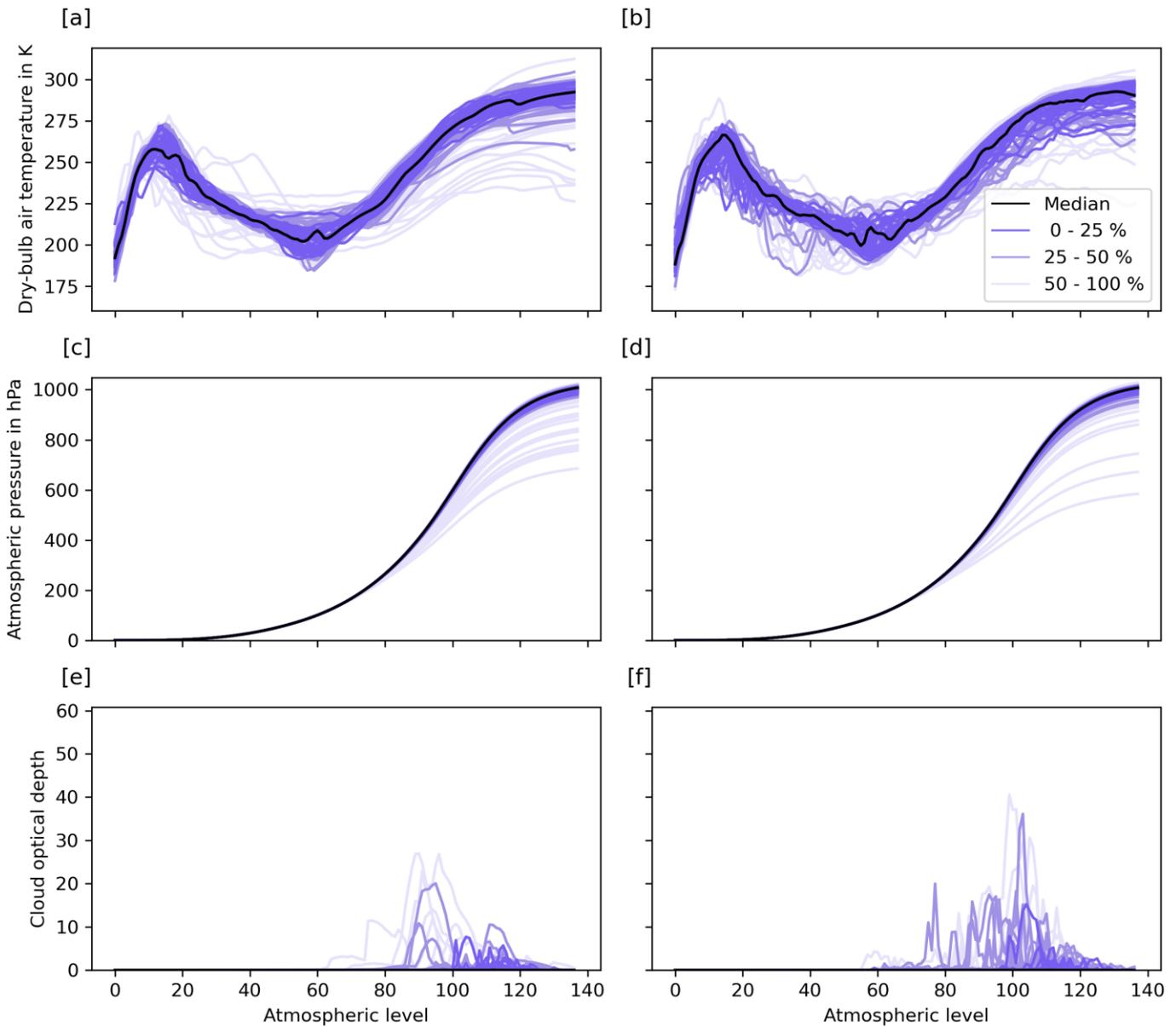

**Figure 4.** Profiles of (**a**, **c**, **e**) real NWP-SAF and (**b**, **d**, **f**) Gaussian-copula-generated data for (**a–b**) dry-bulb air temperature, (**c–d**) atmospheric pressure, and (**e–f**) cloud optical depth. The median profile is shown in black, with a random selection of 90 profiles grouped in batches of 3 (i.e. each having 30 profiles) for the central 0 %–25 %, outer 25 %–50 %, and 50 %–100 % calculated with band depth statistics (López-Pintado and Romo, 2009).

### 3.2 Machine learning

To evaluate whether ML emulators trained on augmented datasets have lower prediction errors compared to the baseline, here we use the test dataset $\mathbf{X}_{test}$ of 10 000 profiles defined in Sect. 2.2. Statistics are computed based on a vector of differences $\boldsymbol{d}$ between the physically predicted baseline $\mathbf{Y}_{test}$ and ML-emulated $\mathbf{Y}'_{test}$ (i.e. $\boldsymbol{d} = \mathbf{Y}_{test} - \mathbf{Y}'_{test}$). From this, the mean bias (MB = $\frac{1}{N}\sum_{i=1}^{N} d_i$) and mean absolute error (MAE = $\frac{1}{N}\sum_{i=1}^{N} |d_i|$) for the set of $N$ profiles are computed. Box plots of MB and MAE are shown in Fig. 5. Summary MB and MAE for the ML emulator with the lowest MAE using an augmentation factor of 10x are reported in Table 2. A qualitative side-by-side comparison of baseline and ML-predicted profiles using Gaussian-copula-generated profiles with an augmentation factor of 10x is shown in Fig. 6.

MBs (Fig. 5a) across all copula types and augmentation factors are generally improved, with median MBs and respective spreads decreasing with larger augmentation factors. Overall, the Gaussian copula model performs better than vine-parametric and vine-nonparametric models. MAEs (Fig. 5b) show a net improvement from the baseline across all copula models and augmentation factors. When using an augmentation factor of 1x (i.e. with double the amount of training data), the median MAE is reduced to approximately 1.1 W m$^{-2}$ from a baseline of approximately 1.4 W m$^{-2}$ and further reduced with increasing augmentation factors. In the best case, corresponding to an augmentation factor of 10x (i.e. with an additional 100 000 synthetic profiles), the copula and ML emulator combinations with the lowest MAE (Table 2) show that MBs are reduced from a baseline of 0.08 W m$^{-2}$ to -0.02 and -0.05 W m$^{-2}$ for Gaussian and vine-nonparametric, respectively, but increased to 0.10 W m$^{-2}$ for vine-parametric. MAEs are reduced from a baseline of 1.17 W m$^{-2}$ to 0.45, 0.56, and 0.44 W m$^{-2}$ for Gaussian, vine-parametric, and vine-nonparametric copula types, respectively.

The ML training configuration with the lowest overall MB and MAE combination during inference corresponds to a Gaussian copula and augmentation factor of 10x (Table 2). Errors between the physically predicted $\mathbf{Y}_{test}$ and ML-predicted $\mathbf{Y}'_{test}$ are shown for the baseline (Fig. 6a) and Gaussian copula case (Fig. 6b). These are shown grouped by their central 0 %–25 %, outer 25 %–50 %, and 50 %–100 %. Qualitatively, most ML-generated profiles show improvements. The most central 25 % profiles are within ±20 W m$^{-2}$ for the Gaussian copula case and about ±40 W m$^{-2}$ for the baseline case. Near-surface errors (levels 130-BOA) are reduced to approximately ±5 W m$^{-2}$ from approximately ±10 W m$^{-2}$.

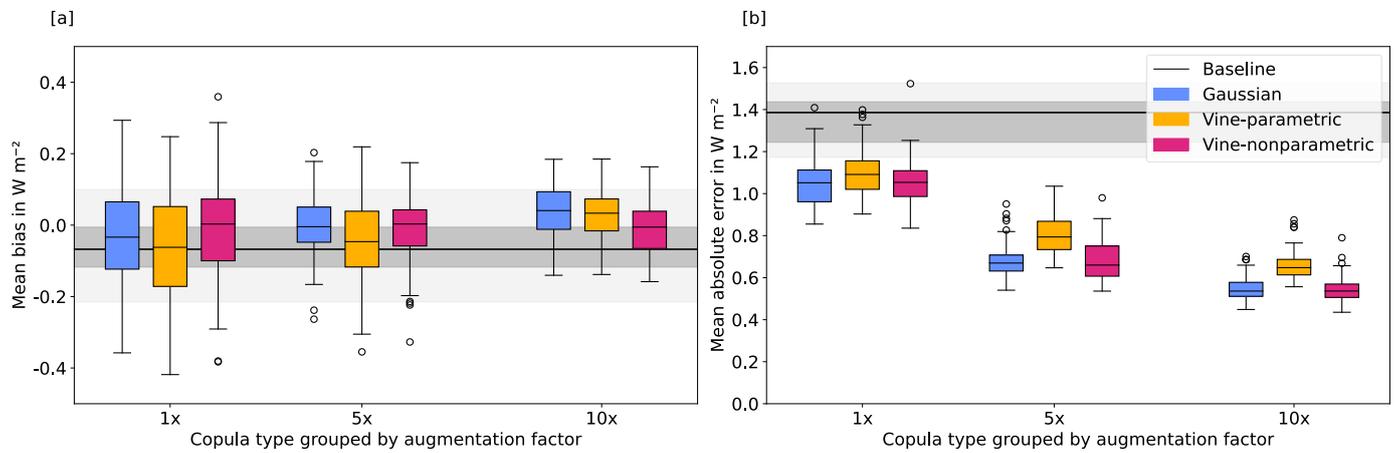

**Figure 5.** Errors grouped by different copula types (Gaussian: blue, vine-parametric: yellow, vine-nonparametric: red) and augmentation factors (1x, 5x, 10x) for the mean bias (MB; **a**) and mean absolute error (MAE; **b**). The median for the baseline case is shown in black, and the range is shaded in grey.

**Table 2.** Mean bias (MB) and mean absolute error (MAE) for baseline and copula cases. Statistics shown for the ML emulator combination with the lowest MAE. Baseline trained using 10 000 real NWP-SAF profiles. Copula cases were trained using 110 000 profiles (10 000 real and 100 000 synthetic), i.e. with an augmentation factor of 10x. Bold indicates the lowest error.

| Case | MB (W m$^{-2}$) | MAE (W m$^{-2}$) |
| --- | --- | --- |
| Baseline | 0.08 | 1.17 |
| Gaussian | **-0.02** | 0.45 |
| Vine-parametric | 0.10 | 0.56 |
| Vine-nonparametric | -0.05 | **0.44** |

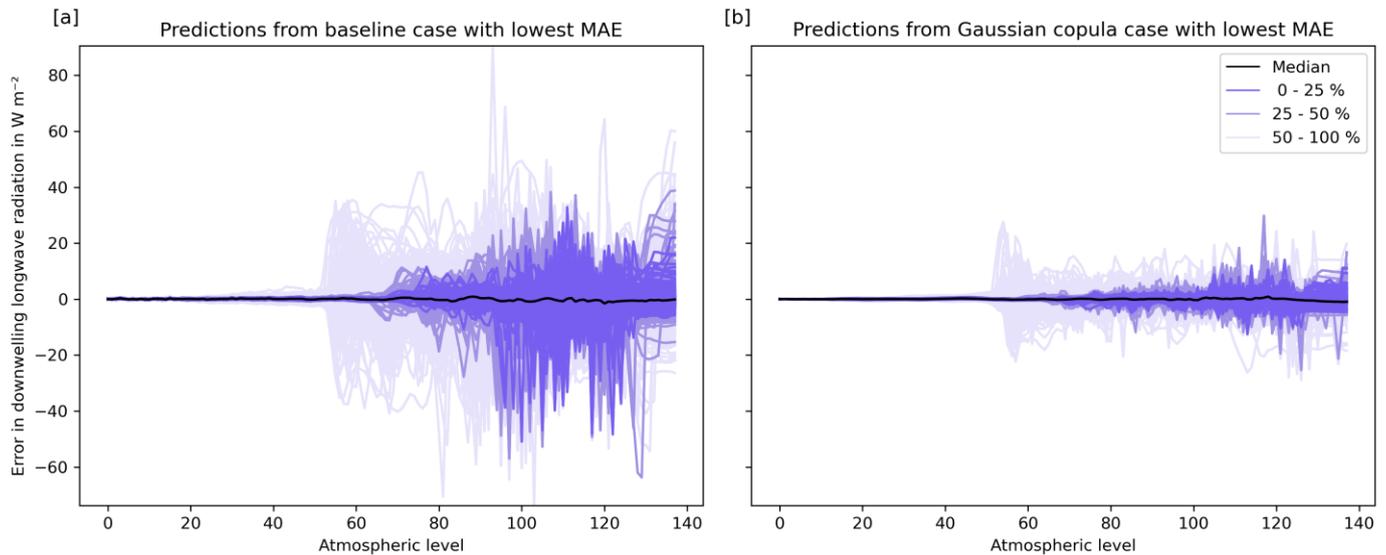

**Figure 6.** Prediction errors for (**a**) baseline and (**b**) data-augmented emulator using 110 000 profiles (10x augmentation factor; Gaussian copula). The median (most central) profile is shown in black, and the most central 25 %, outer 25 %–50 %, and 50 %–100% profiles are computed using band depth statistics and shown in shades of blue.

## 4  Discussion and conclusion

Results from a qualitative comparison of synthetically generated profiles (Fig. 4) show that synthetic profiles tend to be less smooth and noisier than the real NWP-SAF. Nevertheless, a machine-learning evaluation shows that errors for emulators trained with augmented datasets are cut by up to 75% for the mean bias (from 0.08 to -0.02 W m$^{-2}$; Table 2) and by up to 62% for the mean absolute error (from 1.17 to 0.44 W m$^{-2}$; Table 2).

In this study, we show how copula-based models may be used to improve the prediction of ML emulators by generating augmented datasets containing statistically similar profiles in terms of their individual behaviour and dependence across variables (e.g. dry-bulb air temperature at a specific level and across several levels). Although the focus of this paper is to evaluate copula-based data generation models to improve predictions of ML emulators, we speculate that the same or similar methods of data generation have the potential to be used in several other ML-related applications, such as to (i) test architectures (e.g. instead of cross-validation, one may generate synthetic datasets of different size to test the effect of sample size on different ML architectures), (ii) generate data for un-encountered conditions (e.g. for climate change scenarios by extending data ranges or relaxing marginal distributions), and (iii) compress data (e.g. by storing reduced parameterized versions of the data if the number of samples is much larger than the number of features).

Although so far, we have only highlighted the main benefits of copula-based models, several limiting factors should also be considered. A key factor for very high-dimensional data is that both Gaussian and vine copula models scale quadratically in the number of features – in terms of both memory and computational complexity. This can be alleviated by imposing structural constraints on the model, for example using structured covariance matrix or truncating the vine after a fixed number of trees. However, this limits their flexibility and adds some arbitrariness to the modelling process. A second drawback compared to GANs is that the model architecture cannot be tailored to a specific problem, like images. For such cases, a preliminary data compression step as in Tagasovska et al. (2019) may be necessary.

As highlighted here, data augmentation for ML emulators may be of particular interest to scientists and practitioners looking to achieve a better generalization of their ML emulators (i.e. synthetic data may act as a regularizer to reduce overfitting; Shorten &

Khoshgoftaar, 2019). Although a comprehensive analysis of prediction errors using different ML architectures is out of scope, our work is a first step towards further research in this area. Moreover, although we did not explore the generation of data for unencountered conditions (e.g. by extending the range of air temperature profiles while keeping a meaningful dependency across other quantities and levels), the use of copula-based synthetic data generation may prove useful to make emulators more resistant to outliers (e.g. in climate change scenario settings) and should be investigated in future research.

**Code and data availability**

Software, data, and tools are archived with a Singularity (Kurtzer et al., 2017) image and deposited on Zenodo as described in the scientific reproducibility section of Meyer et al. (2020). Users may download the archive (Meyer, 2021) and optionally re-run experiments.

**Author contribution**

Conceptualization, D.M.; Data curation, D.M.; Formal analysis, D.M., T.N.; Investigation, D.M.; Methodology, D.M., T.N., R.H.; Software, D.M.; Resources, D.M.; Validation, D.M.; Visualization, D.M.; Writing – original draft preparation, D.M., T.N.; Writing – review & editing, D.M., T.N., R.H..

**Acknowledgments**

We thank three anonymous reviewers for their useful comments and feedback.